\documentclass{article}
\usepackage[utf8]{inputenc}
\usepackage{latexsym}
\usepackage[normalem]{ulem}

\usepackage{authblk}
\usepackage[top=0.8in, bottom=1in, left=0.85in, right=0.85in]{geometry}
 
\usepackage[outline]{contour}
\usepackage{comment}
\usepackage{graphicx}
\usepackage{color}
\usepackage{amsmath, amsthm, amssymb}
\usepackage{enumerate}
\usepackage{float}
\usepackage{url}
\usepackage{stackrel}
\usepackage{mathrsfs,dsfont}
\usepackage{caption}
\usepackage{wrapfig}
\usepackage{bbm}
\usepackage{bm}
\usepackage{epigraph}
\usepackage{tikz-qtree}
\usepackage{tikz}
\usepackage{indentfirst}
\usepackage{float}
\usepackage{mathtools}
\usepackage{subfigure}
\usepackage{tabularx}
\usepackage{booktabs}

\usepackage{lipsum}

\usepackage{abstract}

\newcommand\blfootnote[1]{%
  \begingroup
  \renewcommand\thefootnote{}\footnote{#1}%
  \addtocounter{footnote}{-1}%
  \endgroup
}

\newcommand{\RNum}[1]
{\uppercase\expandafter{\romannumeral #1\relax}}

\newcommand{\tb}{\textcolor{black}}

\title{\tb{Enhancing Bandwidth Efficiency for Video Motion Transfer Applications using Deep Learning Based Keypoint Prediction}}

\author[1]{Xue Bai}
\author[1]{Tasmiah Haque}
\author[2]{Sumit Mohan}
\author[3]{Yuliang Cai}
\author[4]{Byungheon Jeong}
\author[1]{Ádám Halász}
\author[1,*]{Srinjoy Das}
\affil[1]{West Virginia University, Morgantown, WV, 26506, USA}
\affil[2]{Intel Corporation, Santa Clara, CA, USA }
\affil[3]{University of Southern California, Los Angeles, CA, 90089, USA }
\affil[4]{Coupa Software, San Mateo, CA, USA}

\date{ }

\begin{document}

\maketitle

\blfootnote{*Corresponding author: srinjoy.das@mail.wvu.edu (Srinjoy Das)}
\blfootnote{Email: xb0002@mix.wvu.edu,\  th00027@mix.wvu.edu,\ sumit.mohan@intel.com,\ caiyulia@usc.edu,\ joseph.jeong@coupa.com, }
\blfootnote{halasz@math.wvu.edu,\ srinjoy.das@mail.wvu.edu.}

\textbf{Abstract:} 
We propose a \tb{deep learning based}
novel prediction framework for \tb{enhanced} bandwidth reduction in motion transfer enabled video applications such as \tb{video conferencing}, 
virtual reality gaming and privacy preservation for patient health monitoring. To model complex motion, we use the First Order Motion Model (FOMM) that represents dynamic objects using learned keypoints along with their local affine transformations. Keypoints are extracted by a self-supervised keypoint detector and organized in a time series corresponding to \tb{the} video frames. Prediction of keypoints,
\tb{to enable transmission using lower frames per second on the source device,}
is performed using a Variational Recurrent Neural Network (VRNN). The predicted keypoints are then synthesized to video frames using an optical flow estimator and a generator network. \tb{This} efficacy of leveraging keypoint based representations in conjunction with VRNN based prediction for both \tb{video} animation and reconstruction is demonstrated on three diverse datasets. For real-time applications, our results \tb{show} the effectiveness of our proposed architecture \tb{by} enabling up to \tb{2x} additional bandwidth reduction over existing keypoint based \tb{video motion transfer} frameworks without significantly compromising video quality.


\section{Introduction}
In today's era, characterized by an increasing demand for ubiquitous services, telecommuting, immersive visual interactions and realistic simulations, there is a proliferation of applications such as video conferencing, augmented reality (AR), virtual reality (VR) gaming and remote medical monitoring \cite{videoconference,ar,vr,medicalprivacy}. Video motion transfer is a transformative technique which involves the transposition of motion from one context to another that has been proposed to efficiently implement such applications, and can be used to enhance user experiences as well as generate creative expressions. Keypoint based representations for the source image and driving videos have been previously used to enable motion transfer based applications \cite{monkeynet,fommpaper}. Such architectures have been more successful in achieving high video quality while enabling a high degree of compression for bandwidth efficiency in comparison to traditional video codec based methods. Some examples of applications where real-time motion transfer can play a critical role are as described below:
\begin{itemize}
\item \textbf{Video Conferencing on Mobile Devices:} Implementing motion transfer with keypoints generated in real-time on devices such as cellphones enables efficient bandwidth utilization, particularly for streaming applications, live interactions, and video conferencing. This can ensure a fluid and high-quality user experience while conserving network resources.
\item \textbf{Virtual Reality (VR) Gaming:}
In VR gaming, real-time motion transfer from human players to animated characters ensures precise and immediate in-game responses to users' physical actions, contributing to a more authentic and enjoyable gaming experience.

\item \textbf{Medical Monitoring Privacy:}
In the context of medical applications, where the care provider is often located remotely, maintaining patient privacy is of paramount concern. In such cases real-time motion transfer can be used to transmit critical patient information such as body movements without disclosing the identity of the subject.
\end{itemize}
In this paper we propose using Deep Learning based keypoint prediction 
in conjunction with  motion transfer pipelines that use unsupervised generation of keypoints for performing video synthesis. 
We demonstrate our approach using the First Order Motion Model (FOMM) \cite{fommpaper} to represent complex motion, augmented with time series prediction via Variational Neural Networks (VRNN) \cite{vrnn}. \tb{In comparison to existing keypoint based video motion transfer schemes, our architecture enables a higher bandwidth reduction for transmission, and lowers the compute requirements of resource constrained client devices such as mobile phones or AR/VR devices used in such applications.  This results in a net 20x or higher bandwidth reduction, as the savings from our prediction framework are additive to the 10x or more bandwidth savings, that can be realized from existing video motion transfer schemes} \cite{videoconference}.


\section{Related Works}

Video prediction and motion transfer have been discussed and studied
from several perspectives using
a broad range of approaches. 
Some recent methods related to these areas are discussed in this section.

\textbf{Video prediction in pixel and high level feature spaces}:
In order to predict and synthesize videos using object dynamics at pixel level, several feature learning strategies have been developed using techniques such as adversarial training \cite{adverserial} and gradient difference loss functions \cite{gdl}. An autoencoder style convolutional neural network has been used in \cite{temporal} to learn pixel-wise long-term dependencies from time-lapse videos, whereas in \cite{multifrequency} a multi-frequency analysis that decomposes images into different frequency bands has been utilized to deal with distortion and lack of temporal coherence. These prediction methods accumulate a large error due to variability in consecutive frames. In \cite{convolutionPixel}, pixels are generated using convolution with input patches by applying a predicted kernel which enables capturing spatio-temporal contextual information. However, such approaches still face challenges \tb{for tasks involving} future prediction. Additionally, forecasting of small and slow-moving objects \tb{is often not accurate using pixel based methods.} 
To overcome the problems of prediction in the high dimensional pixel space, several representation learning methods have been explored 
such as semantic segmentation, human pose estimation and using extracted keypoint coordinates. Before the advent of neural networks, pose parameters or Principal Component Analysis (PCA) \cite{PCA1,PCA2} were used to reconstruct a video sequence using only the face boundaries. However applications of pose-guided prediction methods \cite{humanpose,pose1,pose2,pose3} have been limited to only those videos which include human presence. In semantic segmentation, semantic elements obtained from visual scenes are utilized instead of pixel level representations \cite{semantic}. The prediction process is decomposed \cite{semantic2} into current frame segmentation and future optical flow prediction in conjunction with LSTM (Long-Short term Memory) \cite{lstm} based temporal modeling. On the other hand, keypoint based 
representations contain particulars at the object level which produces better results for trajectory prediction and action recognition. In \cite{unsupervisedkp} future video frames are reconstructed using keypoints and a reference frame, thereby preventing the accumulation of errors in the pixel space by focusing on the dynamics within the keypoint coordinate space.
For our work we employ the concept of keypoint prediction for both video reconstruction and
animation using multiple datasets.

\textbf{Video motion transfer:} 
Motivated by the requirement of making high quality \tb{interactive and immersive
video calling}
feasible for users with poor connectivity or limited data plans, keypoint based motion transfer models which enable reconstruction of \tb{objects} in real-time have been developed in order to realize sufficient bandwidth savings. The Neural Talking Heads algorithm \cite{talkinghead} initially applies meta-learning on a large video dataset and afterwards becomes capable of generating talking head models from just a few or even a single image of a person using adversarial training.
With the help of a stream of landmarks/keypoints and an initial face embedding, the face of one person and the movement of another person can be integrated. The limitation of this approach is the necessity of landmark adaptation where  the model underperforms if landmarks are used from a different person.
A model using keypoint representations for synthesizing talking head videos using a source image for appearance and a driving video for motion has been proposed in \cite{videoconference}. This enables video conferencing with no perceptible change in video quality while providing significant reduction for the required bandwidth.
Monkey-Net proposed in \cite{monkeynet}, is known as the first object-agnostic deep learning based model for \tb{video} animation, since here the source image is animated based on the keypoint movements detected in the driving video that are learned in a self-supervised manner.
The principal drawback is that Monkey-Net relies on a simplistic zeroth order keypoint model which results in poor generation quality when there are significant changes in the object's pose. To address this challenge, the First Order Motion Model (FOMM) \cite{fommpaper} has been proposed that decouples appearances and motion information \tb{of video frames} with the help of local affine transformations and coordinates of self learned keypoints. 
A novel end-to-end unsupervised framework for motion transfer using keypoints has been recently proposed in \cite{spline} in order to address challenges posed by significant pose differences between source and driving images in \tb{video} animation. This thin-plate spline motion estimation is utilized to make the optical flow more adaptable, as well as provide enhanced restoration of features in missing areas using multi-resolution occlusion masks which leads to the generation of high-quality images.
Our current work which utilizes the FOMM motion transfer pipeline is a significant extension and improvement of our earlier results reported in \cite{yuliang} using Monkey-Net.  Our method is quite general and can be applied for realizing bandwidth savings in other keypoint based architectures such as \cite{videoconference,spline} which are used for tasks such as video reconstruction and motion transfer.

\section{The Proposed Pipeline}
\begin{figure}[H]
  \centering
\includegraphics[width=0.85\textwidth, height = 0.4\textwidth]{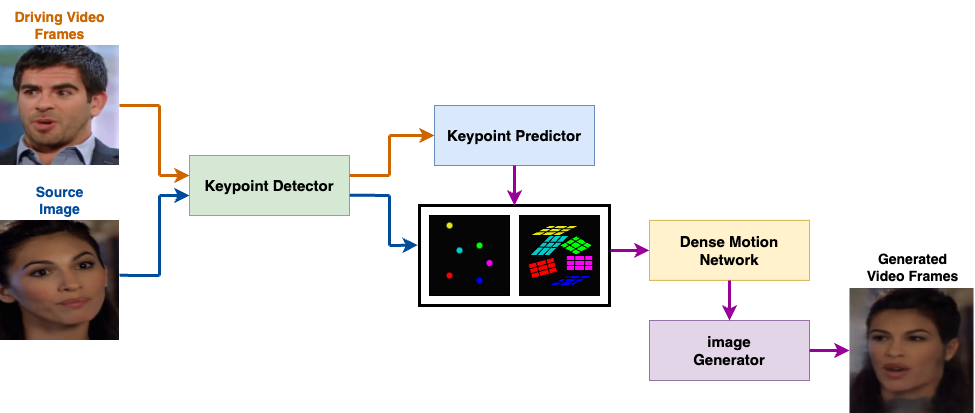}
  \caption{Components of our proposed pipeline for keypoint prediction and video synthesis} \label{fig:3}
\end{figure}

We apply keypoint prediction within the First Order Motion Model (FOMM) motion transfer pipeline \cite{fommpaper}. \tb{The architecture of our pipeline, enabling keypoint prediction and video synthesis, is illustrated in Figure \ref{fig:3}.} Inputs to the pipeline consist of a source image $\mathbf{S}$ and a sequence of driving video frames $\mathbf{D}$. In this architecture, keypoints are first extracted in an unsupervised manner by a keypoint detector \cite{kpdetector} for both the source image $\mathbf{S}$ and the driving video frames $\mathbf{D}$. The extracted keypoints are represented as keypoint locations and the parameters of local affine transformations, which model the motion around each individual keypoint. Using the affine transformations helps the FOMM capture more complex motions as compared to earlier models like Monkey-Net \cite{monkeynet}. Following this \tb{detection} network, forecasting of keypoints is performed using a Variational Recurrent Neural Network (VRNN) which has been shown to effectively deal with complex underlying dynamics of high dimensional time series data \cite{vrnn}. For comparison purposes we also perform forecasting \tb{of keypoints} using a Recurrent Neural Network (RNN) \cite{rnn} and a Variational Autoencoder (VAE) \cite{vae}. Following prediction of keypoints, a dense motion network is used to align the feature maps from the source image $\mathbf{S}$ to the driving video frames $\mathbf{D}$. Finally a generator module is used to produce a video of the source object as per the dynamics in the driving video. 

\newpage
In this work we use 10 keypoints for every video frame, each of which consists of coordinates $x,y$. In addition, a dense motion field is used to align the keypoints captured for the source frame $\mathbf{S}$ and those captured for the object motions in driving frames $\mathbf{D}$. In general this correspondence can be \tb{described} by a function $\mathcal{T}_{S \leftarrow D}$. For simplicity of modeling\tb{,} this function is linearized around each keypoint in which case it suffices to capture the 2 x 2 Jacobian matrix of the transformation $\mathcal{T}_{S \leftarrow D}$ which estimates the motion in the vicinity of each keypoint. Therefore the overall time series of keypoints for every video frame consists of 20 coordinates of the keypoints and 40 components from the Jacobian matrix. In the following subsections, we briefly discuss the underlying framework for the prediction of keypoint time series using the RNN, VAE and VRNN. 

\subsection{Recurrent Neural Network (RNN)}
Keypoint prediction can be performed using a Recurrent Neural Network (RNN)\cite{rnn} which is designed to handle sequential data $ \bm{x} = (\bm{x}_1, \bm{x}_2, \ldots, \bm{x}_T)$ by introducing a hidden state $\bm{h}$. 
At a given timestep $t$, 
the RNN updates its hidden state $\bm{h}_t$ based on input $\bm{x}_t$ as below:
\begin{align}
    \bm{h}_t = f_{\eta}(\bm{x}_t, \bm{h}_{t-1})
\end{align}
Here $f$ is a nonlinear activation function and $\eta$ is a set of learnable parameters of $f$. During prediction the RNN generates its output as below:
\begin{align}
    p(\bm{x}_{t+1} | \bm{x}_{<=t}) = g_{\tau}(\bm{h}_{t})
\end{align}
where $g$ is a nonlinear function with learnable parameters $\tau$ that maps the hidden state $\bm{h}_{t}$ to a probability distribution $p$ of outputs $\bm{x}_{t+1}$. For our work given keypoints $\bm{x}_1, \bm{x}_2, \ldots \bm{x}_T$ an RNN can be trained by minimizing the mean squared error loss between its predicted outputs and their corresponding ground truth values.

\subsection{Variational Autoencoder (VAE)}
Since the keypoints form a high-dimensional dataset, we can also consider performing prediction using the latent space of a generative model such as a Variational Autoencoder (VAE) \cite{vae}. In a VAE at a given timestep $t$, the input $\bm{x}_t$ is initially fed into a probabilistic encoder $q$ parameterized by weights $\phi$. This is used to estimate the posterior distribution $q_{\phi}(\bm{z}_t|\bm{x}_t)$ which is chosen to be a Gaussian as below: 
\begin{align}
    q_{\phi}(\bm{z}_t|\bm{x}_t) = \mathcal{N}(\mu_\phi(\bm{x}_t), \sigma^2_\phi(\bm{x}_t))
\end{align}
Here $\bm{z}_t$ is the latent variable and $\mu_\phi(\bm{x}_t)$, $\sigma^2_\phi(\bm{x}_t)$ are the mean and variance of the Gaussian distribution respectively. The variable $\bm{z}_t$ is sampled using the reparameterization trick \cite{repara} as follows:
\begin{align}
    \bm{z}_t = \mu_\phi(\bm{x}_t) + \sigma_\phi(\bm{x}_t) \odot \epsilon
\end{align}
where $\epsilon$ is a sample from a standard normal distribution, $\epsilon \sim \mathcal{N}(0,I)$ and $\odot$ denotes the element-wise product. The decoder parameterized by weights $\theta$ generates the reconstructed data $\bm{x}_t$ given a sampled latent variable $\bm{z}_t$ as below:
\begin{align}
    p_{\theta}(\bm{x}_t|\bm{z}_t) = \mathcal{N}(\mu_\theta(\bm{z}_t), \sigma_\theta^2(\bm{z}_t))
\end{align}
where $p_\theta(\bm{x}_t|\bm{z}_t)$ is the likelihood of generating the data $\bm{x}_t$ given the latent variable $\bm{z}_t$. The VAE can be trained by minimizing the loss function as below:
\begin{align}
    \mathcal{L}_{VAE}
    = - \mathbb{E}_{q_\phi(\bm{z}_t|\bm{x}_t)}[log \ p_\theta(\bm{x}_t|\bm{z}_t)] + KL[q_\phi(\bm{z}_t|\bm{x}_t) || p(\bm{z}_t)]
\end{align}
Here $\theta$ and $\phi$ are the parameters of the decoder and encoder respectively and $p(z_t)$ is the prior distribution over the latent space which is usually a standard Gaussian distribution. The first term encourages the reconstruction of the input data $x_t$, and the second term is a regularizer which minimizes the Kullback-Leibler (KL) divergence between the prior distribution $p(z)$ over the latent space versus the posterior distribution $q_\phi(z|x)$.

Once trained a VAE can be used to perform reconstruction tasks where an input $x_t$ can be reconstructed as $x'_t$ as the output of the decoder. For our application given keypoints $ \bm{x} = (\bm{x}_1, \bm{x}_2, \ldots, \bm{x}_T)$ the VAE can be trained so that the decoder can generate an output $x_{t+\tau}$ for a given lag $\tau = 1,2,\dots, k$ based on input $x_t$ at time $t$. 

\subsection{Variational Recurrent Neural Network (VRNN)}
A Variational Recurrent Neural Network (VRNN) \cite{vrnn} combines the principles of Variational Autoencoders (VAE) with Recurrent Neural Networks (RNN) to model sequential data. The VRNN consists of an RNN that contains a VAE at each time step. 
\begin{figure}[H]
  \centering
\includegraphics[width=0.5\textwidth, height = 0.35\textwidth]{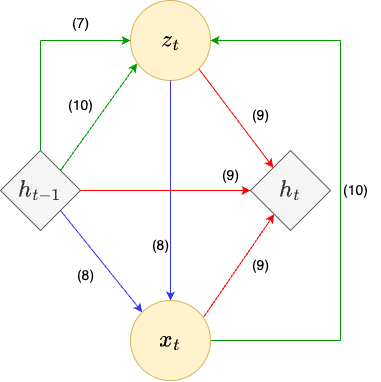}
  \caption{Graphical representation of VRNN describing the dependencies between the variables in Eqs. (7)-(10). The green arrows correspond to the computations involving the (conditional) prior and posterior on $\bm z_t$. The blue arrows show the computations involving the generative network. The computations for $\bm h_t$ are shown with red arrows.} \label{fig:2}
\end{figure}
Unlike a regular VAE, the prior $z_t$ in a VRNN depends on the preceding inputs through ${\bm h}_{t-1}$ and is assumed to follow a Gaussian distribution with parameters $\bm{\mu}_{0,t}, \bm{\sigma}^2_{0,t}$ as below: 

\begin{align}
    p(\bm{z}_t) = \mathcal{N}(\bm{\mu}_{0,t}, \mathrm{diag}(\bm{\sigma}^2_{0,t})), \text{where} \  [\bm{\mu}_{0,t}, \bm{\sigma}_{0,t}] = \varphi^{prior}_{\tau}(\bm{h}_{t-1})
    \label{eq7}
\end{align} 
The generating distribution of the data $\bm{x}_t$ parameterized by $\bm{\mu}_{\bm{x},t}, \bm{\sigma}^2_{\bm{x},t}$ is conditioned on the latent variable $\bm{z}_t$ and the hidden state $\bm{h}_{t-1}$ as follows:
\begin{align}
    p(\bm{x}_t | \bm{z}_t) = \mathcal{N} (\bm{\mu}_{\bm{x},t}, \mathrm{diag}(\bm{\sigma}^2_{\bm{x},t})), \text{where} \ [\bm{\mu}_{\bm{x},t}, \bm{\sigma}_{\bm{x},t}] = \varphi^{dec}_{\tau}(\varphi^{\bm{z}}_{\tau}(\bm{z}_t), \bm{h}_{t-1})
        \label{eq8}
\end{align}
The hidden state $\bm{h}_t$ of the RNN is updated using the recurrence equation: 
\begin{align}
    \bm{h}_t = f(\varphi^{\bm{x}}_{\tau}(\bm{x}_t), \varphi^{\bm{z}}_{\tau}(\bm{z}_t), \bm{h}_{t-1}; \theta)
        \label{eq9}
\end{align}
where $f$ is a non-linear activation function.
The approximate posterior $q(\bm{z}_t|\bm{x}_t)$ parameterized by ${\bm \mu}_{{z},t}, {\bm \sigma^2}_{\bm{z},t}$ is conditioned on both $\bm{x}_t$ and $\bm{h}_{t-1}$ as below:
\begin{align}
    q(\bm{z}_t|\bm{x}_t) = \mathcal{N}({\bm \mu}_{\bm{z},t}, {\bm \sigma^2}_{\bm{z},t}), \text{where} \ \left[{\bm \mu}_{\bm{z},t}, {\bm \sigma}_{\bm{z},t}\right] = \varphi^{enc}_\tau (\varphi^{\bm{x}}_\tau(\bm{x}_t), \bm{h}_{t-1})
        \label{eq10}
\end{align}
Here $\varphi^{prior}_{\tau}$ in Eq. (\ref{eq7}), $\varphi^{dec}_{\tau}$ in Eq. (\ref{eq8}), $\varphi^{\bm{x}}_{\tau}, \varphi^{\bm{z}}_{\tau}$ in Eqs. (\ref{eq9}) and $\varphi^{enc}_\tau$ in Eq. (\ref{eq10}) all refer to non-linear functions that can be realized using Deep Neural Networks. A graphical representation of these computations for the VRNN is shown in Figure \ref{fig:2} \cite{vrnncite}.

\medskip
\noindent For our application given keypoints $\bm{x}_1, \bm{x}_2, \ldots \bm{x}_T$ the VRNN is trained using a loss function which combines KL divergence with the reconstruction loss as below:
\begin{align}
\mathcal{L}_{VRNN} = 
\mathbb{E}_{q(\bm{z}_{\leq T}|\bm{x}_{\leq T})} \left[\sum^{T}_{t=1}
\left(KL(q(\bm{z}_t|\bm{x}_{\leq t}, \bm{z}_{<t}) || p(\bm{z}_t | \bm{x}_{<t}, \bm{z}_{<t})\right) - log \ p(\bm{x}_t|\bm{z}_{\leq t}, \bm{x}_{<t}) )\right]
\end{align}

\subsection{Training and Inference}
Our proposed pipeline offers two operational modes: reconstruction and transfer. In reconstruction mode, the source image $\mathbf{S}$ and the sequence of driving video frames $\mathbf{D}$ are sourced from the same video. For instance, if video $A$ consists of $k$ frames, the source image $\mathbf{S}$ corresponds to the first frame of video $A$, and the driving video frames encompass frames $2$ through $k$ from video $A$. On the other hand, in transfer mode, the source image $\mathbf{S}$ and the sequence of driving video frames $\mathbf{D}$ originate from different videos. The pipeline is first trained end-to-end for video reconstruction, i.e., without keypoint prediction.
The keypoint detector from the trained model is then utilized to generate keypoints, which are then employed to train the RNN, VAE, and VRNN separately based on their respective loss functions as described earlier. 

During inference in either reconstruction or transfer mode, a source image $\mathbf{S}$ and a sequence of driving video frames $\mathbf{D}$ are input to the keypoint detector during the initial $M$ frames of a video sequence. Subsequently, no inputs are provided for the following $N$ frames. During this phase, the RNN, VAE, and VRNN predict the keypoints, after which the optical flow and generator networks are employed to synthesize the next $N$ video frames. Although our model is trained in reconstruction mode, inference can be carried out in either mode to obtain the generated video sequence. In transfer mode, where there is no ground truth video for reference, we generate this using the FOMM pipeline without keypoint prediction and employ it as the ground truth during prediction.

\section{Experimental Results}
We present \tb{our} experimental results to analyze the performance of the three networks introduced in Section 3 using our proposed keypoint prediction and video synthesis pipeline as shown in Figure \ref{fig:3}.
Our experiments are conducted using \tb{three different networks for prediction}: RNN, VAE, and VRNN across three diverse datasets: Mgif \cite{mgifdata}, Bair \cite{bairdata} and VoxCeleb \cite{voxdata}, in both reconstruction and transfer mode. In the following subsections, we will provide an introduction to the three datasets and present our experimental results. We demonstrate the superior performance of VRNN compared to RNN and VAE across all three datasets in both reconstruction and transfer mode.
\subsection{Datasets}
We assess the quality of video generation using three datasets. 
\begin{itemize}
    \item \textbf{Mgif dataset:} We employ 284 training and 34 testing videos from the Mgif dataset, which captures periodic movements of various animals. The videos exhibit varying frame counts, ranging from a minimum of 168 frames to a maximum of 10,500 frames, capturing the diverse distribution of the data. This dataset enables a comprehensive analysis of the model's capacity to capture a range of animal movements with distinct patterns. 
    \item \textbf{Bair dataset:} We use 5001 training and 256 testing videos from the Bair dataset. Each video contains a fixed set of 30 frames and consists of robotic arms in motion to capture different objects. This dataset is helpful for evaluating the proposed architecture's performance in complex environments characterized by diverse backgrounds and irregular movements. 
    \item \textbf{VoxCeleb dataset:} We utilize 3884 training and 44 testing videos from the VoxCeleb dataset, which contains diverse interview scenes with celebrities. Within the dataset, the number of frames in each video varies, ranging from a minimum of 72 frames to a maximum of 1228 frames. 
    This dataset tests the model's capability to capture subtle facial changes, a crucial aspect for applications such as videoconferencing where conveying delicate expressions is essential. 
\end{itemize}

\subsection{Evaluation Procedures}

We compare the predictive capabilities of RNN, VAE, and VRNN across the Mgif, Bair, and VoxCeleb datasets, considering both reconstruction and transfer modes. For both modes, we take \tb{a} block of $k$ frames (where $k\geq 1$) of ground truth keypoints as input and predict the following block of $k$ frames of keypoints. \tb{We then iterate this} process over the entire video \tb{sequence}
until the remaining number of frames are less than $2k$. \tb{This is done for each video in their respective daatset, during training and inference.}

In reconstruction mode, the network takes the source image $\mathbf{S}$ and driving video frames $\mathbf{D}$ from the same video $A$. 
The Mean Squared Error (MSE) and Frechet Video Distance (FVD) \cite{fvd} between the reconstructed video $A_r$ and the corresponding original ground truth video $A$ are then calculated to evaluate the prediction performance.
\tb{Lower values of MSE and FVD indicate lower amount of degradation versus the corresponding ground truth and therefore superior video quality.}
In transfer mode, the network operates with the source image $\mathbf{S}$ from video $B$ and driving video images $\mathbf{D}$ from video $A$. 
\tb{In this mode, there is no ground truth corresponding to the generated video $B_t$.}
To address this we generate video $B'_t$ by performing inference using the FOMM pipeline without Deep Learning based networks for keypoint prediction. This is then used \tb{in lieu of} the ground truth for evaluating the performance in transfer mode.
\subsection{Results on Mgif dataset}
To evaluate the performance of our prediction using the Mgif dataset, we perform long and short \tb{horizon} predictions in both reconstruction and transfer mode. Specifically, for long \tb{horizon} prediction, we set $k=12$, i.e., we use the first 12 frames of ground truth as input to predict next 12 frames and repeat this process for the full video; while for short \tb{horizon} prediction, we choose $k=6$, i.e., we apply the first 6 frames of ground truth as input to predict the next 6 frames and repeat this for the full video.

The prediction performance of VRNN, RNN and VAE for the Mgif dataset in case of long and short \tb{horizon} scenarios are shown in Tables \ref{table:1} and \ref{table:2} respectively. The results demonstrate that across all cases, VRNN consistently outperforms RNN and VAE in tasks involving both reconstructing ground truth videos and generating motion transfer videos. This superiority is evident through both assessment metrics, MSE and FVD. 
The qualitative results for selected videos are shown in Figure \ref{fig:4}. 

\begin{table}[h!]
\centering
\begin{tabular}{|c||c|c|c||c|c|c|} 
 \multicolumn{7}{c}{\textbf{Predicted results for reconstruction mode}} \\
 \hline
 {\textbf{Prediction type
 }} & \multicolumn{3}{c||}{\textbf{MSE}} & \multicolumn{3}{c|}{\textbf{FVD}} \\
 \cline{2-7}
(\# input frames, \# output frames)  & VRNN & RNN & VAE & VRNN & RNN & VAE\\ 
 \hline
 \textbf{(6, 6)} & \textbf{0.0340} & 0.0392 & 0.0380 & \textbf{7.8469}& 9.1903& 8.8407 \\
 \hline
 \textbf{(12, 12)} & \textbf{0.0338} & 0.0388 & 0.0382& \textbf{7.7835} & 8.8653 & 8.8571 \\  
 \hline
\end{tabular}
\vspace{0.2cm}
\caption{MSE and FVD results of Mgif dataset in reconstruction mode}
\label{table:1}
\end{table}

\begin{table}[h!]
\centering
\begin{tabular}{|c||c|c|c||c|c|c|} 
 \multicolumn{7}{c}{\textbf{Predicted results for transfer mode}} \\
 \hline
 {\textbf{Prediction type
 }} & \multicolumn{3}{c||}{\textbf{MSE}} & \multicolumn{3}{c|}{\textbf{FVD}} \\
 \cline{2-7}
(\# input frames, \# output frames)  & VRNN & RNN & VAE & VRNN & RNN & VAE\\ 
 \hline
 \textbf{(6, 6)} & \textbf{0.0171} & 0.0230 & 0.0215 & \textbf{3.7741}& 5.1614& 4.7917\\
 \hline
 \textbf{(12, 12)} &\textbf{0.0164} & 0.0227 & 0.0215 & \textbf{3.5933} & 5.0911 & 4.7714\\  
 \hline
\end{tabular}
\vspace{0.2cm}
\caption{MSE and FVD results of Mgif dataset in transfer mode}
\label{table:2}
\end{table}

\subsection{Results on Bair dataset}
For the Bair dataset, given that each video consists of 30 frames, we perform both long and short \tb{horizon} predictions by selecting $k=15$ and $k=5$, respectively. The prediction results for the Bair dataset are shown in Tables \ref{table:3} and \ref{table:4}. Observing the outcomes, it is evident that for this dataset characterized by complex robotic motion, VRNN consistently outperforms RNN and VAE across both long and short \tb{horizon} prediction tasks. The qualitative results for selected videos are shown in Figure \ref{fig:5}. 

\begin{table}[h!]
\centering
\begin{tabular}{|c||c|c|c||c|c|c|} 
\multicolumn{7}{c}{\textbf{Predicted results for reconstruction mode}} \\
\hline
 {\textbf{Prediction type
 }} & \multicolumn{3}{c||}{\textbf{MSE}} & \multicolumn{3}{c|}{\textbf{FVD}} \\
 \cline{2-7}
(\# input frames, \# output frames)  & VRNN & RNN & VAE & VRNN & RNN & VAE\\ 
 \hline
 \textbf{(5, 5)} & \textbf{0.0660} & 0.0752 & 0.0749 & \textbf{10.8439}& 13.8028& 13.7585 \\
 \hline
 \textbf{(15, 15)} &\textbf{0.0656} & 0.0732 & 0.0759 & \textbf{10.7086} & 13.1366 & 14.1428 \\ 
 \hline
\end{tabular}
\vspace{0.2cm}
\caption{MSE and FVD results of Bair dataset in reconstruction mode}
\label{table:3}
\end{table}

\newpage

\begin{table}[h!]
\centering
\begin{tabular}{|c||c|c|c||c|c|c|} 
 \multicolumn{7}{c}{\textbf{Predicted results for transfer mode}} \\
 \hline
 {\textbf{Prediction type
 }} & \multicolumn{3}{c||}{\textbf{MSE}} & \multicolumn{3}{c|}{\textbf{FVD}} \\
 \cline{2-7}
(\# input frames, \# output frames)  & VRNN & RNN & VAE & VRNN & RNN & VAE\\ 
 \hline
 \textbf{(5, 5)} & \textbf{0.0133} &0.0305 & 0.0267 & \textbf{2.3951}& 6.7901& 5.8773 \\
 \hline
 \textbf{(15, 15)} &\textbf{0.0128} & 0.0231 & 0.0263 & \textbf{2.2500} & 4.8392 & 5.7437\\  
 \hline
\end{tabular}
\vspace{0.2cm}
\caption{MSE and FVD results of Bair dataset in transfer mode}
\label{table:4}
\end{table}

\subsection{Results on VoxCeleb dataset}
\begin{figure}[H]
  \centering
\includegraphics[width=0.9\textwidth, height = 0.54\textwidth]{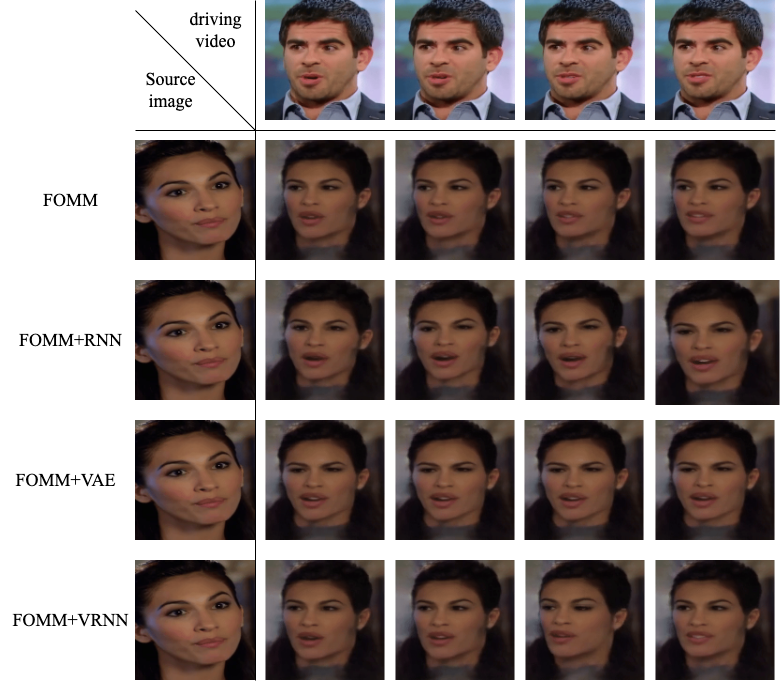}
  \caption{Qualitative results for the VoxCeleb dataset in reconstruction mode (upper panel) and transfer mode (lower panel). In each panel, consecutive frames generated using only FOMM are shown in the second row, and FOMM with keypoints prediction using RNN, VAE, and VRNN are shown in the third, fourth and fifth rows respectively. For reconstruction mode, the first row serves as the ground truth whereas for transfer mode the second row serves as the ground truth.}\label{fig:6}
\end{figure}

For the VoxCeleb dataset, similar to the Mgif dataset, we conduct both long and short \tb{horizon} prediction experiments with $k=12$ and $k=6$, respectively. The prediction results for the VoxCeleb dataset are shown in Table \ref{table:5} and \ref{table:6}. We notice that, even in this scenario involving a more intricate dataset with diverse scenes, VRNN based prediction consistently exhibits superior performance compared to RNN and VAE. The qualitative results for selected videos are shown in Figure \ref{fig:6}.


\begin{table}[h!]
\centering
\begin{tabular}{|c||c|c|c||c|c|c|} 
 \multicolumn{7}{c}{\textbf{Predicted results for reconstruction mode}} \\
 \hline
 {\textbf{Prediction type
 }} & \multicolumn{3}{c||}{\textbf{MSE}} & \multicolumn{3}{c|}{\textbf{FVD}} \\
 \cline{2-7}
(\# input frames, \# output frames)  & VRNN & RNN & VAE & VRNN & RNN & VAE\\ 
 \hline
 \textbf{(6, 6)} & \textbf{0.0483} & 0.0564 & 0.0602 & \textbf{3.7120}& 4.6684& 5.3585 \\
 \hline
 \textbf{(12, 12)} &\textbf{0.0471} & 0.0649 & 0.0651 & \textbf{3.5917} & 6.1379 & 6.2279\\ 
 \hline
\end{tabular}
\vspace{0.2cm}
\caption{MSE and FVD results of VoxCeleb dataset in reconstruction mode}
\label{table:5}
\end{table}

\begin{table}[h!]
\centering
\begin{tabular}{|c||c|c|c||c|c|c|}
 \multicolumn{7}{c}{\textbf{Predicted results for transfer mode}} \\
  \hline
 {\textbf{Prediction type
 }} & \multicolumn{3}{c||}{\textbf{MSE}} & \multicolumn{3}{c|}{\textbf{FVD}} \\
 \cline{2-7}
(\# input frames, \# output frames)  & VRNN & RNN & VAE & VRNN & RNN & VAE\\ 
 \hline
 \textbf{(6, 6)} & \textbf{0.0097} &0.0202& 0.0245& \textbf{0.2325}& 0.8705& 1.2542\\
 \hline
 \textbf{(12, 12)} &\textbf{0.0078} & 0.0305& 0.0292& \textbf{0.1535} & 1.8252 & 1.7347 \\ 
 \hline
\end{tabular}
\vspace{0.2cm}
\caption{MSE and FVD results of VoxCeleb dataset in transfer mode}
\label{table:6}
\end{table}

\begin{figure}[H]
  \centering
\includegraphics[width=0.9\textwidth, height = 0.55\textwidth]{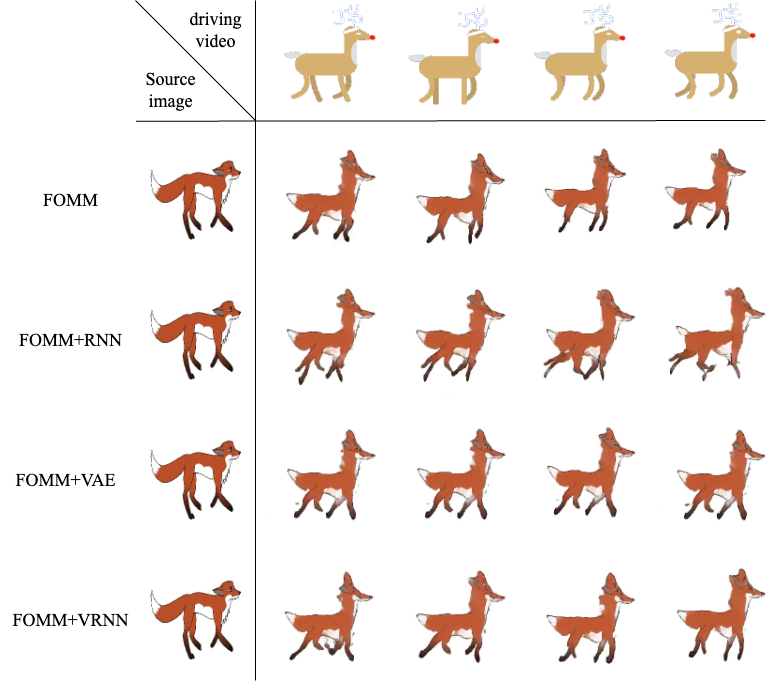}
  \caption{Qualitative results for the Mgif dataset in reconstruction mode (upper panel) and transfer mode (lower panel). In each panel, consecutive frames generated using only FOMM are shown in the second row, and FOMM with keypoints prediction using RNN, VAE, and VRNN are shown in the third, fourth and fifth rows respectively. For reconstruction mode, the first row serves as the ground truth whereas for transfer mode the second row serves as the ground truth.}\label{fig:4}
\end{figure}

\begin{figure}[H]
  \centering
\includegraphics[width=0.9\textwidth, height = 0.54\textwidth]{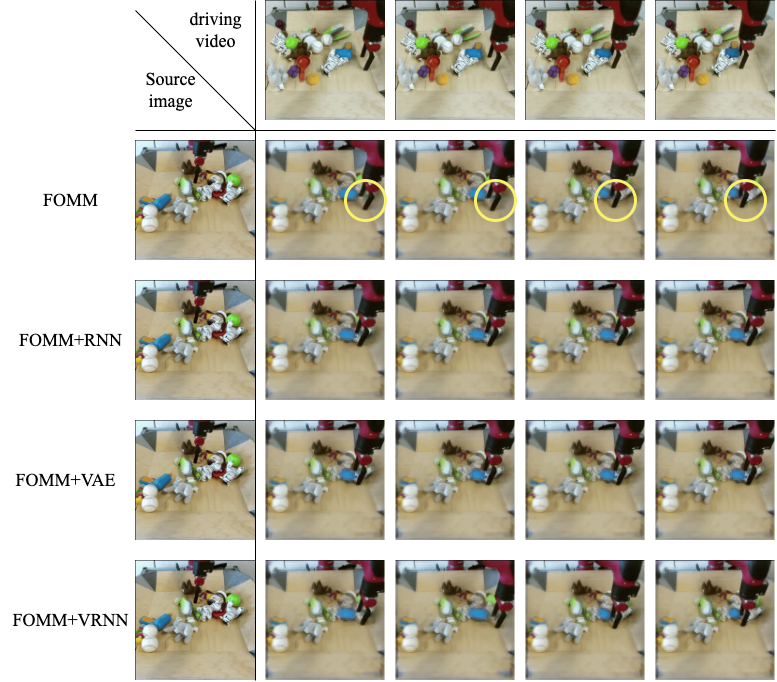}
\caption{Qualitative results for the Bair dataset in reconstruction mode (upper panel) and transfer mode (lower panel). In each panel, consecutive frames generated using only FOMM are shown in the second row, and FOMM with keypoints prediction using RNN, VAE, and VRNN are shown in the third, fourth and fifth rows respectively. For reconstruction mode, the first row serves as the ground truth whereas for transfer mode the second row serves as the ground truth. The circles in both figures are examples of regions where VRNN performs better than RNN and VAE.}\label{fig:5}
\end{figure}

In this work we use two error metrics namely the MSE which focuses primarily on pixel to pixel differences and FVD which is a more comprehensive measure of the difference in the distribution of feature vectors between generated and reference videos. Based on both of these metrics and the qualitative results it can be seen that the outputs generated from VRNN consistently demonstrate superior video quality and temporal coherence. 
\tb{Morever, our qualitative results in Figures \ref{fig:4}, \ref{fig:5}, \ref{fig:6} also demonstrate that using VRNN based predicted keypoints does not significantly affect the quality of the output videos as compared to the baseline case when the full set of keypoints are generated using the FOMM only.}
This supports our \tb{contention} that predicting keypoints with VRNN in the FOMM motion transfer pipeline can be used to generate significant bandwidth savings without seriously compromising video quality.

\newpage

\begin{figure}[H]
  \centering
\includegraphics[width=0.9\textwidth, height = 0.55\textwidth]{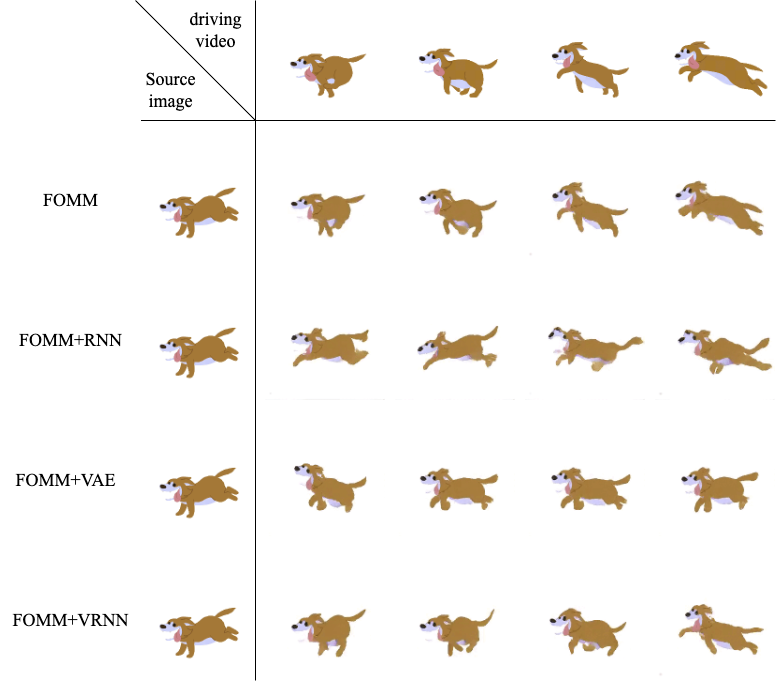}
\end{figure}
\vspace{-0.5cm}

\newpage
\begin{figure}[H]
  \centering
\includegraphics[width=0.9\textwidth, height = 0.54\textwidth]{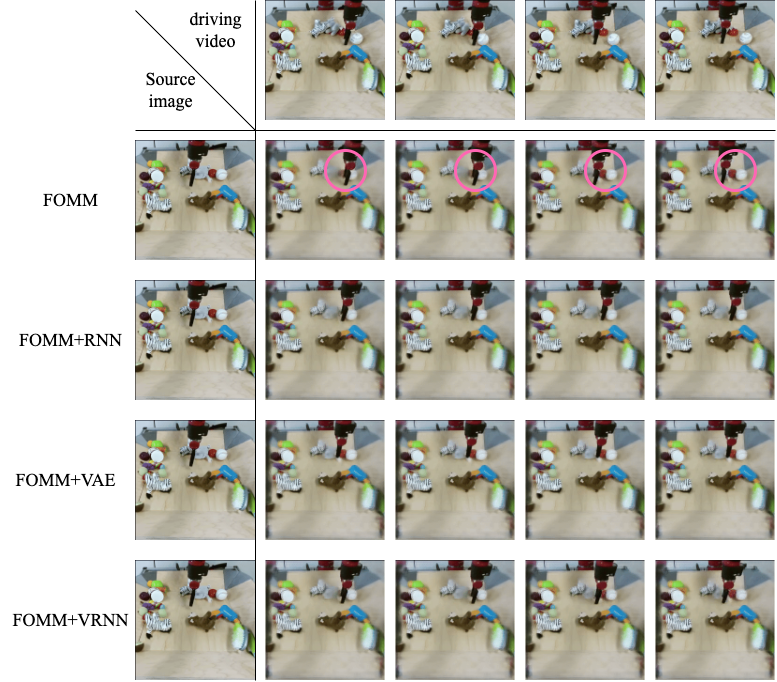}
\end{figure}
\vspace{-0.5cm}

\newpage
\begin{figure}[H]
  \centering
\includegraphics[width=0.9\textwidth, height = 0.54\textwidth]{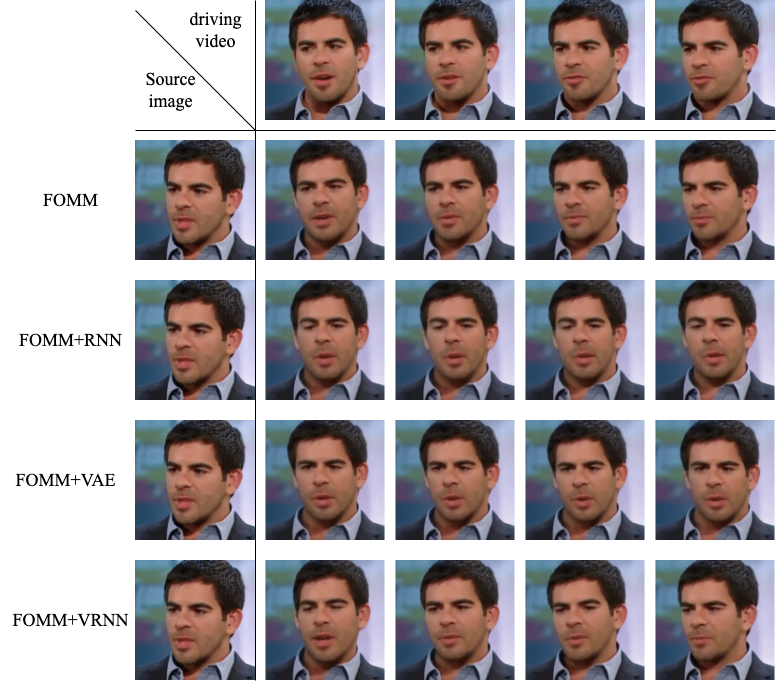}
\end{figure}
\vspace{-0.5cm}
\newpage
\section{Conclusions and Future Directions}

In this paper, we propose a novel approach for \tb{enhancing} bandwidth savings in real-time motion transfer based applications such as video conferencing, \tb{virtual reality gaming} and patient monitoring privacy. This is realized by using a VRNN which seamlessly integrates the advantages of an RNN \tb{that} efficiently captures temporal dependence and a VAE which can effectively perform feature extraction using a probabilistic latent space. By employing representative datasets and varying prediction \tb{horizons}, our empirical analysis shows the efficacy of our proposed architecture in achieving a \tb{2x additional} reduction in bandwidth requirements when compared to cases where prediction is omitted. \tb{This is done by replacing half of the frames using their predicted counterparts from the VRNN.}

Across all datasets, our results consistently demonstrate the superior performance of VRNN in video prediction. 
Specifically, the Mean Squared Error, reflecting pixel-to-pixel differences, and the Frechet Video Distance, which considers both spatial and temporal aspects of videos, exhibit smaller values when compared to those obtained with RNN and VAE based methods. 
As it was pointed out in the original work by Chung et al. (2015) \cite{vrnn} who introduced VRNN for natural speech, 
this type of network is well-suited for modeling sequential data characterized by a complex structure, high intrinsic variability and a high signal-to-noise ratio. The typical application of VRNN is described as time series that exhibit high and sudden variations that are {\em not} due to noise. This is an accurate characterization for keypoints, which follow trajectories that are not deterministic or completely predictable but are highly constrained by geometry and are not significantly affected by position uncertainty. RNNs have a limited ability to deal with sudden variations and randomness in the data that is not due to noise. 
While VAEs are effective in learning probabilistic representations and generating diverse samples, they may face problems in capturing the temporal dependencies and sudden variations present in the data. 
Therefore, the incorporation of structured output functions in VRNNs, coupled with their ability to represent complex non-linear data, positions them as a superior choice for the type of applications that is the focus of this work.
Our future research will focus on using methods such as Transformers \cite{transformer} for  keypoint forecasting \tb{over longer horizons}
\tb{to realize a higher degree of bandwidth savings}
for motion transfer applications.

\subsection*{Acknowledgements}
\noindent X.B. was supported for this work by an Intel grant. The authors would like to acknowledge the Pacific Research Platform, NSF Project ACI-1541349, and Larry Smarr (PI, Calit2 at UCSD) for providing the computing infrastructure used in this project.

\subsection*{Samples of Generated Videos}
\noindent Selected videos from reconstruction and transfer are provided at: \\
https://github.com/xuebai95/Motion-Transfer-Keypoints-Prediction-Videos


\end{document}